# Enhancement Without Contrast: Stability-Aware Multicenter Machine Learning for Glioma MRI Imaging

Sajad Amiri[1], Shahram Taeb[2], Sara Gharibi[1], Setareh Dehghanfard[3], Somayeh Sadat Mehrnia[4], Mehrdad Oveisi[5,6], Ilker Hacihaliloglu[7,8], Arman Rahmim[7,9], Mohammad R. Salmanpour[6,7,9*]

[1]Department of Medicine, Tehran University of Medical Science, Tehran, Iran
[2]Department of Radiology, School of Paramedical Sciences, Guilan University of Medical Sciences, Rasht, Iran
[3]Department of Electrical and Computer Engineering, University of Tehran, Tehran, Iran
[4]Department of Integrative Oncology, Breast Cancer Research Center, Motamed Cancer Institute, ACECR, Tehran, Iran
[5]Department of Computer Science, University of British Columbia, Vancouver, BC, Canada
[6]Technological Virtual Collaboration Company (TECVICO CORP.), Vancouver, BC, Canada
[7]Department of Radiology, University of British Columbia, Vancouver, BC, Canada
[8]Department of Medicine, University of British Columbia, Vancouver, BC, Canada
[9]Department of Basic and Translational Research, BC Cancer Research Institute, Vancouver, BC, Canada

**(\*) Corresponding Author:** Mohammad R. Salmanpour, PhD; msalman@bccrc.ca

## Abstract

**Objective:** Gadolinium-based contrast agents (GBCAs) are essential in glioma imaging but raise safety, cost, and accessibility concerns. Predicting contrast enhancement from non-contrast MRI using machine learning (ML) offers a safer, cost-effective alternative, as enhancement reflects tumor aggressiveness and guides planning. However, variability across scanners and populations limits robust model selection. To address this, we present a stability-aware framework to identify reproducible ML pipelines for predicting glioma MRI contrast enhancement across multicenter cohorts.
**Methods:** A total of 1,446 glioma cases from four TCIA datasets (UCSF-PDGM, UPENN-GB, BRATS-Africa, BRATS-TCGA-LGG) were analyzed. Non-contrast T1WI served as input, with contrast enhancement status derived from paired post-contrast T1WI. 108 Radiomics features were extracted using PyRadiomics following IBSI standards. Forty-eight dimensionality reduction algorithms and 25 classifiers were systematically combined into 1,200 pipelines. Rotational validation was performed by training on three datasets and externally testing on the fourth, repeated across three rotations. Model evaluation incorporated five-fold cross-validation, external testing, and a composite scoring system penalizing instability via standard deviation integration.
**Results:** Cross-validation prediction accuracies ranged from 0.91–0.96, with external testing achieving 0.87 (UCSF-PDGM), 0.98 (UPENN-GB), and 0.95 (BRATS-Africa), averaging~0.93. F1, precision, and recall were stable (0.87–0.96), while ROC-AUC varied more widely (0.50–0.82), reflecting cohort heterogeneity. The MI+ETr pipeline consistently ranked highest, balancing accuracy and stability.
**Conclusion:** This framework demonstrates that stability-aware model selection enables reliable prediction of contrast enhancement from non-contrast glioma MRI, reducing reliance on GBCAs and improving generalizability across centers. It provides a scalable template for reproducible ML in neuro-oncology and beyond.

## 1. Introduction

Magnetic resonance imaging (MRI) is a critically utilized modality in neuro-oncology, as other anatomical techniques do not match its precision in outlining glioma boundaries, peritumoral oedema, or in assessing treatment response [1]. As a result, almost all patients diagnosed with a diffuse glioma, regardless of whether categorized as low-grade (I–II) or high-grade (III–IV), undergo multiple MRI scans from the point of diagnosis through to end-of-life care [2]. Gliomas are indeed a heterogeneous group of primary brain tumors arising from glial cells. The global incidence rate of 6–8 per 100,000 individuals highlights their relative rarity but significant clinical impact due to their aggressive nature and poor prognosis. The fact that about 30-40% of these gliomas are high-grade underscores the burden of malignancy within this tumor category [3] [4]. The diagnostic utility of MRI in neuro-oncology has been significantly enhanced through the routine use of gadolinium-based contrast agents (GBCAs).

GBCAs exploit the paramagnetic properties of Gadolinium ($Gd^{3+}$) to reduce T1-weighted MRI images (T1WI) relaxation time, thereby enhancing the visualization of blood-brain barrier (BBB) disruptions—a feature observed in approximately 80% of high-grade gliomas [5]. This enhancement on post-contrast T1WI provides critical biological and clinical insights, aiding in surgical resection planning, radiotherapy target definition, and the application of Response Assessment in Neuro-Oncology criteria (KJR Online) [6]. Bright (hyperintense) areas typically indicate BBB compromise due to leakage from neovascularization, characteristics of anaplastic astrocytoma, glioblastoma (GBM), and other high-grade lesions. In contrast, the absence of enhancement suggests an intact BBB, as seen in most grade I–II gliomas—though exceptions exist, such as "non-enhancing GBM" or focal enhancement in low-grade tumors [7]. These imaging patterns hold significant prognostic value: enhancement often correlates with aggressive histology and poorer outcomes, guides surgical and radiotherapy targeting, determines eligibility for anti-angiogenic trials, and serves as a key imaging marker for distinguishing true progression from pseudo-progression or pseudo-response during long-term monitoring [8].

Nonetheless, the growing dependence on GBCAs is still under investigation. The use of linear chelates has caused an increase in nephrogenic systemic fibrosis among individuals with kidney issues, leading to black-box warnings and withdrawal of several products by the University of Maryland Medical System [5] [9]. Further autopsy and biopsy studies have revealed that $Gd^{3+}$ accumulates in a dose-dependent fashion in the dentate nucleus, globus pallidus, bone, and skin, even in those with normal renal function [10]. Although a direct clinical syndrome has not been conclusively linked, these discoveries have increased medicolegal scrutiny and patient concern, particularly among children and young adults who need ongoing imaging throughout their lives [9]. Beyond these safety and retention concerns, the widespread reliance on GBCAs also imposes significant economic and logistical burdens, particularly in resource-limited settings [11].

The financial and operational challenges associated with GBCAs have exacerbated accessibility issues in neuroimaging. By 2024, the global market for CT/MRI contrast media will have surpassed USD 6 billion, with projections reaching approximately USD 10 billion by 2030, driven largely by oncology demand. In low- and middle-income countries, the cost of a single GBCA vial can exceed a household's weekly income, while the logistical burdens of intravenous cannulation, creation of testing, and prolonged scan durations further strain limited radiology resources [12]. Compounding these challenges is a critical scientific gap: conventional non-contrast MRI lacks sufficient sensitivity to detect subtle BBB disruptions, while advanced techniques such as diffusion- or perfusion-weighted imaging improve detection but remain imperfect standalone alternatives [13]. Consequently, there is a pressing need for tools that can predict or replicate contrast enhancement (CE) without $Gd^{3+}$ administration.

Emerging computational approaches aim to address this unmet need [14] [15] [16] [17]. Radiomics features (RFs), which provide quantitative descriptors of tumor texture, shape, and intensity from standard MR images, have shown promise in brain disease studies [18] [19] [20]. Single-center investigations report that machine learning (ML) models can distinguish enhancing from non-enhancing gliomas with accuracies exceeding 0.80, though these findings are limited by small sample sizes, heterogeneous preprocessing methods, and insufficient external validation [21] [22]. Meanwhile, deep learning techniques—particularly convolutional neural networks and generative adversarial networks—have explored image-to-image translation to synthesize "virtual contrast" images from pre-contrast scans, achieving structural similarity indices of 0.80–0.90 in early trials [23]. Despite these advances, concerns persist regarding model generalizability, clinical interpretability, and regulatory acceptance. To date, no framework has successfully bridged the transparency of radiomics with the high performance of deep learning to meet the demands of radiologists, oncologists, and regulatory bodies.

A major challenge in ML for medical imaging is robust model selection when a large number of algorithms are available [24]. With dozens of dimension reduction strategies and classifiers, the search space quickly expands to thousands of possible combinations, making it difficult to identify which models will generalize well beyond the training set [25] [26]. This problem becomes even more critical in multicenter cohorts, where differences in scanners, acquisition protocols, and patient demographics can cause models that perform well internally to fail when applied to external data [27]. Traditional approaches often rely on single-center cross-validation, which risks overfitting and inflates performance estimates. Therefore, the key issue is not only achieving high accuracy but also ensuring stability and reproducibility across heterogeneous datasets, requiring systematic, fair, and scalable evaluation frameworks for model selection [28] [29].

This paper presents a comprehensive ML framework for glioma outcome prediction across multicenter datasets. Using 1,446 cases from four large The Cancer Imaging Archive (TCIA) cohorts, the study applies rigorous preprocessing, expert labeling, and standardized radiomics feature extraction to ensure reliable inputs. A large-scale model search was conducted by evaluating 1,200 combinations of 48-dimensional reduction algorithms and 25 classifiers. To address variability and enhance generalizability, the pipeline incorporates rotational dataset partitioning, internal cross-validation, and independent external testing. Model performances are assessed with multiple metrics and integrated into a composite scoring system that balances accuracy and stability, enabling systematic ranking and selection of the best models. The results highlight robust top-performing Data Reduction Augmentation – Classification Algorithm (DRA–CA) pairs, demonstrate reproducibility across folds and external cohorts, and provide a scalable strategy for fair, interpretable, and clinically relevant model selection in glioma radiomics.

## 2. Materials and Methods

i) **Patient Data Preparation**. Four publicly available large glioma datasets comprising a total of 1446 cases from TCIA were analyzed, each containing T1WI with expert-validated tumor segmentations. As illustrated in Table 1, the datasets included 496 samples from UCSF PDGM [30] (299 males, 202 females; mean age 56.87±15.02 years), 660 samples from UPENN-GB [31] (405 males, 266 females; mean age 62.46±12.37 years), 231 samples from BRATS Africa [32], and 59 samples from BRATS TCGA LGG [33]. Imaging datasets from TCIA exhibited variations in scanner types, MRI sequences, acquisition parameters, preprocessing methods, and data quality. For example, the UCSF-PDGM dataset includes preoperative 3T MRI scans with 3D and 2D sequences, processed with eddy current correction, DTI processing, and skull stripping using deep learning. This single-center dataset contains preoperative scans with no prior treatment (except biopsy), some variability in contrast agents, but no missing sequences. The UPENN-GBM dataset features multi-parametric MRI scans from GBM patients, acquired on 1.5T and 3T scanners, with preprocessing including skull-stripping, co-registration, automated tumor segmentation, and radiomic feature extraction. This single-center dataset supports radiogenomic studies and includes only preoperative scans with no missing sequences. The BraTS-Africa dataset includes multiparametric MRI scans from brain tumor patients acquired across six Nigerian centers using 1.5T scanners, with preprocessing steps like N4 bias field correction, skull-stripping, and rigid registration. It contains preoperative scans with variability in scanner types but no missing sequences, supporting diagnostic tool development for African populations. The BraTS-TCGA-LGG dataset offers preoperative multi-parametric MRI scans from glioma patients, acquired across multiple institutions. Preprocessing included skull-stripping, co-registration, and automated tumor segmentation, followed by manual corrections. This dataset supports molecular and outcome studies and also has no missing sequences.

**Table 1.** Summary of TCIA Glioma Datasets: Demographics, Preprocessing, and Partitioning; abbreviations: T1: T1-weighted MRI sequence. T1-CE / T1-Gd: T1-weighted sequence with Contrast Enhancement. "Gd"; Gadolinium, T2: T2-weighted MRI sequence. T2-FLAIR: T2-weighted Fluid-Attenuated Inversion Recovery sequence. DWI: Diffusion-Weighted Imaging. PWI: Perfusion-Weighted Imaging, DTI: Diffusion Tensor Imaging. GBM: Glioblastoma, HGG: High-Grade Glioma, LGG: Low-Grade Glioma,

| Dataset | Subjects | Males | Females | Tumor Grades | MRI Modalities | Survival Data | Access Restrictions |
|---|---|---|---|---|---|---|---|
| BraTS-Africa | 146 | N/A | N/A | LGG/GBM/HGG | T1, T1-CE, T2, T2-FLAIR | Not specified | Public (CC BY 4.0) |
| UCSF-PDGM | 495 | 298 | 203 | Grade II, III, IV | T2, T2/FLAIR, DWI, T1 -Gd | Not specified | Public (CC BY 4.0) |
| BraTS TCGA LGG | 59 | N/A | N/A | LGG (Grades I-II) | T1, T1-Gd, T2, T2-FLAIR | Not specified | Partial Restrictions (TCIA) |
| UPENN-GBM | 630 | 405 | 266 | GBM (Grade IV) | T1, T1-CE, T2, T2/FLAIR, DWI, DSC, and DTI | Overal Survival | Public (CC BY 4.0) |

ii) **Labeling:** The dataset was meticulously labeled to ensure accurate representation of glioma CE status. Image labeling followed a structured workflow: (a) non-contrast T1WI served as the input for RF extraction, and (b) the corresponding contrast-enhanced T1WI was used as the ground truth (Figure 1). Radiologists identified enhancement patterns (e.g., ring-like, nodular) by comparing (a) and (b). Contrast enhancement is a key indicator in glioma imaging, as it reflects BBB disruption, a hallmark of tumor aggressiveness. This feature is critical for refining tumor grading, informing surgical and radiotherapy planning, and guiding treatment monitoring under the RANO criteria. Furthermore, enhancement patterns are crucial for differentiating true progression from pseudoprogression following chemoradiotherapy, a challenge that can otherwise result in premature therapy changes or unnecessary interventions. Given these factors, the ability to reliably predict enhancement, without gadolinium, offers a safer, cost-effective approach while retaining the diagnostic and prognostic value traditionally obtained from contrast imaging. Therefore, binary classification labels were assigned: enhanced = 1, non-enhanced = 0. All labels were independently verified by expert radiologists to preserve clinical relevance and ensure consistency across multicenter data. Moreover, all tumor segmentations were performed by an experienced radiologist and independently validated by a second expert to minimize inter-observer variability. This step ensured the reliability of the regions of interest (ROIs) used for RF extraction.

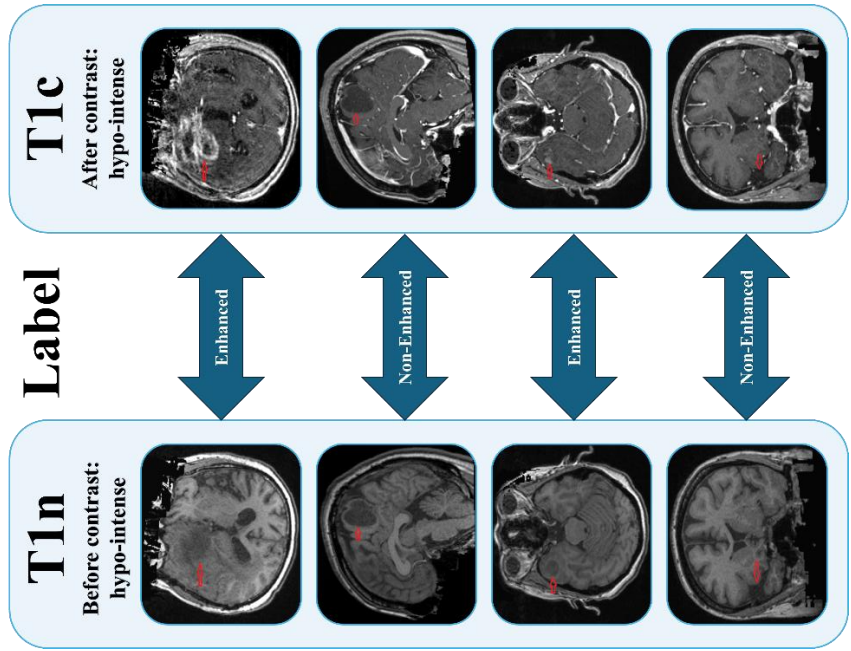

**Fig. 1.** Labeling workflow: non-contrast T1WI (a) used for feature extraction, contrast-enhanced T1WI (b) as ground truth for generation of radiologist-verified binary labels (enhanced = 1, non-enhanced = 0), to be predicted from non-contrast imaging.

iii) **MRI intensity Normalization:** Non-contrast T1WI data were standardized using Min-Max normalization to account for variations in scanner protocols and imaging conditions across multicenter cohorts. This preprocessing step enhanced the comparability of RFs extracted from different datasets.

iv) **RF Extraction:** A comprehensive set of RFs was extracted using PyRadiomics [34], standardized in reference to the image biomarker standardization initiative (IBSI). In total, 108 standardized RFs were considered as the reference set, comprising 19 first-order (FO) features, 15 shape-based features (SF), 23 gray-level co-occurrence matrix (GLCM) features, 16 gray-level size zone matrix (GLSZM) features, 16 gray-level run length matrix (GLRLM) features, 5 neighborhood gray-tone difference matrix (NGTDM) features, and 14 gray-level dependence matrix (GLDM) features.

v) **Rotational Data Partitioning into Training and Test Sets:** We performed a rotational ML analysis in which three datasets were combined and used for five-fold cross-validation, while the remaining dataset was reserved for external testing. This process was repeated three times to ensure robustness. The TCGA-LGG dataset (59 patients) was included only in the five-fold cross-validation due to its insufficiently balanced labels and was therefore excluded from the rotational external testing procedure. Internal validation was performed and optimized via stratified 5-fold cross-validation and grid search.

vi) **Min-Max Normalization of RFs:** Extracted RFs underwent Min-Max normalization to scale values between 0 and 1. This step ensured uniformity in feature ranges, preventing bias in ML models due to varying magnitudes. We used only the four training folds of the cross-validation process for data normalization.

vii) **Dimension Reduction:** A total of 48 dimensionality reduction techniques (25 feature selection algorithms (FSA) and 23 attribute extraction algorithms (AEAs)) are evaluated for their ability to isolate the most informative and non-redundant features. DRAs were configured to reduce the feature space to 10 dimensions. The 25 FSAs encompass three major categories. Filter-based methods include the Chi-Square Test (CST), Correlation Coefficient (CC), Mutual Information (MI), and Information Gain Ratio, which score features independently of classifiers. Statistical tests such as ANOVA F-Test (AFT), ANOVA P-value selection (APT), Chi2 P-value, and Variance Thresholding (VT) are used to assess feature discriminativeness. Wrapper-based methods include Recursive Feature Elimination (RFE), Univariate Feature Selection (UFS), Sequential Forward Selection (SFS), and Sequential Backward Selection (SBS), which iteratively evaluate subsets using model performance. Embedded methods such as Lasso, Elastic Net, Embedded Elastic Net, and Stability Selection perform selection during training. Ensemble-based approaches, including Feature Importance by RandF (FIRF), Extra Trees Importance, and Permutation Importance (Perm-Imp), capture complex non-linear relationships. Additional statistical control methods like False Discovery Rate (FDR), Family-Wise Error (FWE), and multicollinearity handling techniques like Variance Inflation Factor (VIF) are also employed. Finally, dictionary-based strategies use Principal Component Analysis (PCA) or sparse loadings for stability and interpretability.

AEAs offer a complementary strategy by transforming the feature space into lower-dimensional subspaces. The 23 AEAs include linear techniques such as PCA, Truncated PCA, Sparse PCA (SPCA), and Kernel PCA (with RBF and polynomial kernels), which identify uncorrelated projections of maximum variance. Independent Component Analysis (ICA) and its variant FastICA extract statistically independent latent variables. Factor Analysis uncovers hidden structure behind observed features, while Non-negative Matrix Factorization (NMF) yields interpretable parts-based decompositions. SL techniques like Linear Discriminant Analysis (LDA) maximize class separation in the transformed space. Advanced manifold learning techniques capture nonlinear structures and include t-distributed Stochastic Neighbor Embedding (t-SNE), Uniform Manifold Approximation and Projection (UMAP), Isomap, Locally Linear Embedding (LLE), Spectral Embedding, Multidimensional Scaling (MDS), and Diffusion Maps. These are especially useful for visualizing complex relationships in high-dimensional radiomics space. Deep learning methods, such as shallow and deep autoencoders, enable data-driven feature compression through reconstruction optimization. Other strategies include Feature Agglomeration for hierarchical grouping, Truncated SVD (TSVD) for matrix decomposition, and projection-based techniques like Gaussian Random Projection, Sparse Random Projection, and Feature Hashing, offering scalable compression.

Each reduced feature set—whether selected by FSA or derived by AEA—is evaluated using a diverse set of 25 classifiers (CAs). These include linear models such as Tree-based classifiers encompassing Decision Trees, Random Forest (RandF), Extra Trees (ETr), Gradient Boosting (GB), AdaBoost (AB), and HistGradient Boosting (HGB), each utilizing ensemble learning to reduce variance and improve generalization. Meta-ensemble strategies such as Stacking, Voting Classifiers (hard and soft), and Bagging further boosted predictive robustness by aggregating the strengths of multiple base learners. Support Vector Machines (SVM) were implemented with various kernels to handle linear and non-linear classification, while k-Nearest Neighbors (KNN) provided a distance-based, instance-level approach. Several Naive Bayes variants (Gaussian process, Bernoulli, and Complement) were tested for their probabilistic simplicity and computational efficiency. Neural network-based Multi-Layer Perceptron (MLP) facilitated modeling of complex patterns in non-linear spaces, whereas gradient-boosted frameworks such as Light GBM (LGBM) and eXtreme Gradient Boosting (XGB) provided high-performance learning through gradient optimization and feature importance modeling. Additional classifiers included LDA, Nearest Centroid, Decision Stump, Dummy Classifier (DC), Gaussian Process Classifier (GP), and Stochastic Gradient Descent Classifier (SGDC) for diverse modeling strategies.

viii) **Rotational Model Selection.** We propose a comprehensive model evaluation and selection pipeline for learning tasks in multicenter studies. The pipeline incorporates three-fold rotational validation and five-fold internal cross-validation within each rotation, ensuring a robust assessment of both performance and stability across combinations of DRAs and classifiers. Performance metrics are computed at multiple levels and aggregated through a composite scoring system, enabling systematic model ranking and selection.

*Multi-Rotation and Cross-Validation Scheme.* To account for center-related variability and enhance generalizability, the entire training and validation process is repeated across three rotation splits. Each rotation simulates a distinct configuration of data partitioning across sites, thereby mimicking real-world deployment scenarios. Within each rotation:

- Each DRA-classifier pair is evaluated using 5-fold cross-validation.
- Performance metrics are computed in each fold and averaged to derive robust estimates.
- Both internal validation metrics (from 5-fold cross-validation) and external test metrics (from a held-out external set per rotation) are recorded.

This process resulted in three averaged cross-validation and external metric values per metric per model, along with corresponding standard deviations (SD) across the five folds. Importantly, only the 5-fold cross-validation results were used for model scoring and selection. The external test performances were held out entirely during the model ranking process to avoid information leakage.

*Performance Metrics.* We computed the following five evaluation metrics: Accuracy, F1 Score, Precision, Recall, and Area Under the Curve (AUC). For each DRA-classifier pair:

- The mean of each metric is calculated across the 5-fold cross-validation within a rotation (denoted as $\mu_i^{(r)}$ for metric i in rotation r).
- The SD of each metric across folds is also computed (denoted as $\sigma_i^{(r)}$).
- Over 3 rotations, this produces a total of:
  - 15 mean values per model: 3 rotations × 5 metrics
  - 15 SD values per model: 3 rotations × 5 metrics

*Aggregation and Normalization.* To enable fair comparison across different metrics and models, we apply min-max normalization to the metric values before scoring. Let:
- $M_{ij}$ is the average of metric i across folds in rotation j.
- $S_{ij}$ is the SD of metric i across folds in rotation j.

We normalize the metric means (Equation 1) and SD (Equation 2) across all models:

$$\widehat{M_{ij}} = \frac{M_{ij} - \min(M_i)}{\max(M_i) - \min(M_i)} \quad (1)$$

$$\widehat{S_{ij}} = \frac{S_{ij} - \min(S_i)}{\max(S_i) - \min(S_i)} \quad (2)$$

Then we invert the normalized SD to compute a stability score (Equation 3):

$$\textbf{Stability}_{ij} = 1 - \widehat{S_{ij}} \quad (3)$$

*Composite Scoring Formula.* The final model selection score aggregates both accuracy and stability across all metrics and rotations (Equation 4). Specifically, for each DRA-classifier pair, we compute:

$$\text{Final Score} = \frac{1}{2}\sum_{i=1}^{5}\sum_{r=1}^{3}\left(\widehat{M_i^{(r)}} + \left(1 - \widehat{S_i^{(r)}}\right)\right) \quad (4)$$

where:
- 5 metrics × 3 rotations = 15 normalized metric averages
- 5 metrics × 3 rotations = 15 normalized SD (converted to stability)
- Total = 30 terms, and the final score is divided by 20 to normalize it to the range [0, 1] while balancing performance and stability equally.

Model Ranking and Selection. Each model is then:
- Assigned a final score as computed above
- Ranked in descending order of score (higher is better)
- Mapped to its cross-validation and external metrics for interpretability

*Model Ranking and Selection.* Each model is then:
- Assigned a final score as computed above
- Ranked in descending order of score (higher is better)
- Mapped to its cross-validation and external metrics for interpretability

The pipeline is compatible with multicenter designs and large-scale model comparisons, enabling automatic, fair, and interpretable model selection under realistic clinical settings.

## 3. Results

Table 2 presents the Top 10 performing DRA–CA pairs for glioma outcome prediction (enhanced vs. non-enhanced), evaluated through the proposed model evaluation pipeline. The pipeline incorporates three-fold rotational validation, five-fold cross-validation, and composite scoring, allowing for systematic and unbiased ranking of all 1,200 tested combinations generated from 48 DRAs paired with 25 CAs.

Among these, the MI+ETr pair emerged as the top performer with the highest overall score (0.941), showing balanced and stable performance across all major metrics: accuracy (0.94 ± 0.02), F1 score (0.92 ± 0.02), precision (0.94 ± 0.02), and recall (0.93 ± 0.02). This consistent profile underscores its robustness in distinguishing enhanced from non-enhanced glioma cases. Feature Embedding (FEW) + ETr and Extra Trees Importance feature selection (ETIm) + GP followed closely, both achieving accuracies of 0.94 ± 0.03 with strong F1 scores (0.93 ± 0.02 and 0.93 ± 0.01, respectively), demonstrating their ability to balance predictive sensitivity and specificity.

Other high-ranking models, such as AFT+LGBM classifier, Recursive Feature Elimination (RFE) +MLP, and FEW+LGBM, maintained similarly strong accuracies (~0.94) with narrow variability across cross-validation folds, reflecting the pipeline's ability to capture stable performance across multiple multicenter data partitions. Interestingly, ETr emerged repeatedly as a dominant classifier in the top-performing combinations (e.g., MI+ETr, FEW+ETr, UFS+ETr, RFE+ETr, TSVD+ETr), suggesting its strong adaptability when paired with different DRAs.

All Top 10 pairs exhibited relatively small SD across accuracy, F1, precision, and recall, underscoring their reproducibility and resilience to data heterogeneity. While the area under the receiver operating characteristic curve (ROC-AUC) values showed more variability (ranging from 0.77 to 0.82), this did not compromise the overall robustness of the top-ranked models. Taken together, these results highlight that the proposed pipeline reliably identifies models with stable and generalizable performance, positioning them as strong candidates for multicenter glioma outcome prediction tasks. Supplementary File 1, Sheet 1, presents the complete list of average rotational performances, rankings, and scores for all 1,200 CA+DRA combinations. Moreover, no significant differences were observed among the top 10 models based on the paired t-test and the Benjamini–Hochberg false discovery rate correction for five-fold cross-validation results.

**Table 2.** Figure X. Model evaluation pipeline with three-fold rotational validation, five-fold cross-validation. Metrics are aggregated into a composite score for systematic model ranking and selection. The Top 10 performing dimension reduction–classifier pairs for glioma outcome prediction are shown across different datasets. CA: Classification Algorithm, DRA: Dimension Reduction Algorithm ROC-AUC: The area under the receiver operating characteristic curve, MI: Mutual Information, FEW: Feature Embedding, ETIm: Extra Trees Importance, AFT: ANOVA F-Test, RFE: Recursive Feature Elimination, UFS: Univariate Feature Selection, APT: ANOVA P-value selection, TSVD: Truncated Singular Value Decomposition, ETr: Extra Trees Classifier, GP: Gaussian Process, LGBM: Light Gradient Boosting Machine, MLP: Multilayer Perceptron, XGB: eXtreme Gradient Boosting.

| Five-Fold Cross Validation | | | | | | | |
|---|---|---|---|---|---|---|---|
| DRA + CA | Rank | Score | Accuracy | F1 Score | Precision | Recall | ROC-AUC |
| MI+ETr | 1 | 0.941381 | 0.94 ± 0.02 | 0.92 ± 0.02 | 0.94 ± 0.02 | 0.93 ± 0.02 | 0.81 ± 0.10 |
| FEW+ETr | 2 | 0.937860 | 0.94 ± 0.02 | 0.93 ± 0.02 | 0.94 ± 0.02 | 0.93 ± 0.02 | 0.78 ± 0.13 |
| ETIm+GP | 3 | 0.934510 | 0.94 ± 0.03 | 0.93 ± 0.01 | 0.94 ± 0.03 | 0.92 ± 0.03 | 0.82 ± 0.03 |
| AFT+LGBM | 4 | 0.934090 | 0.94 ± 0.03 | 0.93 ± 0.02 | 0.94 ± 0.03 | 0.93 ± 0.02 | 0.79 ± 0.11 |
| RFE_MLP | 5 | 0.933245 | 0.94 ± 0.03 | 0.92 ± 0.02 | 0.94 ± 0.03 | 0.93 ± 0.02 | 0.79 ± 0.14 |
| FEW+LGBM | 6 | 0.933074 | 0.94 ± 0.03 | 0.93 ± 0.02 | 0.94 ± 0.03 | 0.93 ± 0.03 | 0.79 ± 0.12 |
| UFS+ETr | 7 | 0.933029 | 0.94 ± 0.02 | 0.92 ± 0.02 | 0.94 ± 0.02 | 0.93 ± 0.02 | 0.80 ± 0.11 |
| APT+XGB | 8 | 0.932931 | 0.94 ± 0.03 | 0.92 ± 0.02 | 0.94 ± 0.03 | 0.93 ± 0.02 | 0.77 ± 0.14 |
| RFE+ETr | 9 | 0.932844 | 0.94 ± 0.03 | 0.93 ± 0.02 | 0.94 ± 0.03 | 0.93 ± 0.03 | 0.81 ± 0.09 |
| TSVD+ETr | 10 | 0.932762 | 0.94 ± 0.03 | 0.93 ± 0.02 | 0.94 ± 0.03 | 0.92 ± 0.03 | 0.77 ± 0.09 |

External testing results for the Top 10 DRA–CA combinations identified in Table 2 are detailed in Table 3. Consistent with the internal cross-validation findings, the MI+ETr pair remained the best-performing combination, achieving the highest composite score (0.941). Its performance across key metrics was balanced, with an accuracy of 0.93 ± 0.06, an F1 score of 0.91 ± 0.06, a precision of 0.93 ± 0.06, and a recall of 0.91 ± 0.08, demonstrating strong reproducibility of internal results in an unseen cohort. Similarly, FEW+ETr and ETIm+GP maintained competitive performance, each reaching an accuracy of 0.93 ± 0.06 but with slightly reduced F1 scores (0.89 ± 0.12 and 0.88 ± 0.11, respectively). These results suggest that while predictive accuracy remained stable, the balance between precision and recall was more sensitive to variations in external data.

Across the Top 10 models, external testing confirmed a high degree of stability in accuracy, precision, and recall, with mean values consistently around 0.93 and SD tightly bound. This reproducibility across independent datasets highlights the robustness of the pipeline's ranking methodology. However, ROC-AUC values demonstrated greater variability (ranging from 0.50 to 0.77), reflecting the influence of dataset heterogeneity on threshold-dependent metrics. Notably, some models, such as RFE+MLP, achieved relatively high accuracy and precision but showed substantial fluctuations in ROC-AUC (0.50 ± 0.29), indicating that performance consistency across decision thresholds is less reliable in certain DRA-classifier pairings.

When comparing internal cross-validation (Table 2) and external validation (Table 3), a strong alignment in ranking patterns was observed, with MI+ETr, FEW+ETr, and ETIm+GP consistently emerging as top performers. The close match between cross-validation and external testing supports the robustness of the evaluation pipeline and provides confidence that the identified top models are generalizable across multicenter datasets. At the same time, the discrepancies in ROC-AUC highlight areas for further refinement, suggesting that while overall classification performance is stable, the discriminative ability across thresholds warrants closer investigation. Similar to cross-validation, no significant improvement was observed among the top 10 models in external testing.

**Table 3.** Model evaluation pipeline with three-fold rotational validation and held-out external testing. Metrics are aggregated into a composite score for systematic model ranking and selection. The Top 10 performing dimension reduction–classifier pairs for glioma outcome prediction are shown across different datasets. CA: Classification Algorithm, DRA: Dimension Reduction Algorithm, ROC-AUC: The area under the receiver operating characteristic curve, MI: Mutual Information, FEW: Feature Embedding, ETIm: Extra Trees Importance, AFT: ANOVA F-Test, RFE: Recursive Feature Elimination, UFS: Univariate Feature Selection, APT: ANOVA P-value selection, TSVD: Truncated Singular Value Decomposition, ETr: Extra Trees Classifier, GP: Gaussian Process, LGBM: Light Gradient Boosting Machine, MLP: Multilayer Perceptron, XGB: eXtreme Gradient Boosting.

| External Test | | | | | | | |
|---|---|---|---|---|---|---|---|
| **DRA + CA** | **Rank** | **Score** | **Accuracy** | **F1 Score** | **Precision** | **Recall** | **ROC-AUC** |
| **MI+ETr** | 1 | 0.941381 | 0.93 ± 0.06 | 0.91 ± 0.06 | 0.93 ± 0.06 | 0.91 ± 0.08 | 0.70 ± 0.13 |
| **FEW+ETr** | 2 | 0.937860 | 0.93 ± 0.06 | 0.89 ± 0.12 | 0.93 ± 0.06 | 0.91 ± 0.09 | 0.69 ± 0.06 |
| **ETIm+GP** | 3 | 0.934510 | 0.93 ± 0.06 | 0.88 ± 0.11 | 0.93 ± 0.06 | 0.91 ± 0.09 | 0.77 ± 0.10 |
| **AFT+LGBM** | 4 | 0.934090 | 0.93 ± 0.06 | 0.90 ± 0.10 | 0.93 ± 0.06 | 0.91 ± 0.09 | 0.73 ± 0.07 |
| **RFE_MLP** | 5 | 0.933245 | 0.94 ± 0.06 | 0.89 ± 0.12 | 0.94 ± 0.06 | 0.91 ± 0.09 | 0.50 ± 0.29 |
| **FEW+LGBM** | 6 | 0.933074 | 0.93 ± 0.06 | 0.90 ± 0.10 | 0.93 ± 0.06 | 0.91 ± 0.09 | 0.74 ± 0.06 |
| **UFS+ETr** | 7 | 0.933029 | 0.93 ± 0.06 | 0.92 ± 0.05 | 0.93 ± 0.06 | 0.91 ± 0.08 | 0.70 ± 0.12 |
| **APT+XGB** | 8 | 0.932931 | 0.93 ± 0.06 | 0.89 ± 0.12 | 0.93 ± 0.06 | 0.91 ± 0.09 | 0.73 ± 0.06 |
| **RFE+ETr** | 9 | 0.932844 | 0.93 ± 0.06 | 0.91 ± 0.06 | 0.93 ± 0.06 | 0.91 ± 0.08 | 0.70 ± 0.10 |
| **TSVD+ETr** | 10 | 0.932762 | 0.93 ± 0.06 | 0.89 ± 0.12 | 0.93 ± 0.06 | 0.91 ± 0.09 | 0.74 ± 0.05 |

Table 4 summarizes the Top 10 performing DRA–CA pairs for glioma outcome prediction, trained on UPENN-GB, BRATS-Africa, and BRATS-TCGA-LGG, with UCSF-PDGM used exclusively for external testing. During internal cross-validation, all Top 10 models achieved consistently high performance, with mean accuracies of 0.96 and narrow error margins across F1 score, precision, and recall. The highest-ranked pair, MI+ETr, achieved an internal accuracy of 0.96 ± 0.01, F1 score of 0.93 ± 0.01, precision of 0.96 ± 0.01, and recall of 0.95 ± 0.01, with a ROC-AUC of 0.70 ± 0.02. Other top-performing models, including FEW+ETr, ETIm+GP, and AFT+LGBM, showed nearly identical accuracies of 0.96 and balanced performance across metrics, reflecting stable and reproducible results in multicenter cross-validation.

When evaluated on the external UCSF-PDGM dataset, model performance remained strong but with modest reductions compared to internal validation, reflecting the challenge of generalizing across independent cohorts. The top-ranked MI+ETr pair achieved an external accuracy of 0.87 ± 0.00, F1 score of 0.85 ± 0.01, precision of 0.87 ± 0.00, recall of 0.82 ± 0.01, and ROC-AUC of 0.81 ± 0.04. Comparable results were observed for FEW+ETr and ETIm+GP, with external accuracies of 0.86–0.87 and F1 scores of 0.75. These findings demonstrate that while predictive accuracy and precision translated well across datasets, F1 scores and ROC-AUC showed greater variability, suggesting sensitivity to label distribution and class imbalance in external data.

Overall, the strong alignment between internal cross-validation and external testing underscores the robustness and reproducibility of the proposed evaluation pipeline. The repeated appearance of Extra Trees (ETr) as a top-performing classifier in multiple high-ranking pairs highlights its adaptability and stability when paired with different DRAs. Although performance metrics declined modestly on external testing, the maintenance of accuracies near 0.87 across models suggests that the identified top-performing pairs are well-suited for multicenter glioma outcome prediction. Supplementary File 1, Sheet 2, presents the complete list of average five-fold cross-validation and external testing performances, rankings, and scores for all 1,200 CA+DRA combinations.

**Table 4.** The Top 10 performing dimension reduction–classifier (DRA–CA) pairs for glioma outcome prediction, trained on UPENN-GB, BRATS Africa, BRATS TCGA LGG, with UCSF PDGM used exclusively for external testing. ROC-AUC: The area under the receiver operating characteristic curve, MI: Mutual Information, FEW: Feature Embedding, ETIm: Extra Trees Importance, AFT: ANOVA F-Test, RFE: Recursive Feature Elimination, UFS: Univariate Feature Selection, APT: ANOVA P-

| Classifier | Score | Rank | Five-Fold Cross Validation by UCSF PDGM, Africa, BRATS TCGA LGG | | | | | External Test by UPENN-GB | | | | |
|---|---|---|---|---|---|---|---|---|---|---|---|---|
| | | | Accuracy | F1 Score | Precision | Recall | ROC-AUC | Accuracy | F1 Score | Precision | Recall | ROC-AUC |
| MI+ETr | 0.941381 | 1 | 0.96 ± 0.01 | 0.93 ± 0.01 | 0.96 ± 0.01 | 0.95 ± 0.01 | 0.70 ± 0.02 | 0.87 ± 0.00 | 0.85 ± 0.01 | 0.87 ± 0.00 | 0.82 ± 0.01 | 0.81 ± 0.04 |
| FEW+ETr | 0.93786 | 2 | 0.96 ± 0.01 | 0.94 ± 0.02 | 0.96 ± 0.01 | 0.95 ± 0.01 | 0.64 ± 0.02 | 0.86 ± 0.00 | 0.75 ± 0.00 | 0.86 ± 0.00 | 0.80 ± 0.00 | 0.69 ± 0.07 |
| ETIm+GP | 0.93451 | 3 | 0.96 ± 0.00 | 0.93 ± 0.01 | 0.96 ± 0.00 | 0.95 ± 0.00 | 0.79 ± 0.04 | 0.87 ± 0.00 | 0.75 ± 0.00 | 0.87 ± 0.00 | 0.80 ± 0.00 | 0.88 ± 0.01 |
| AFT+LGBM | 0.93409 | 4 | 0.96 ± 0.00 | 0.93 ± 0.01 | 0.96 ± 0.00 | 0.95 ± 0.01 | 0.66 ± 0.04 | 0.87 ± 0.00 | 0.79 ± 0.05 | 0.87 ± 0.00 | 0.81 ± 0.01 | 0.81 ± 0.07 |
| RFE_MLP | 0.933245 | 5 | 0.96 ± 0.00 | 0.93 ± 0.01 | 0.96 ± 0.00 | 0.95 ± 0.00 | 0.62 ± 0.04 | 0.87 ± 0.00 | 0.75 ± 0.00 | 0.87 ± 0.00 | 0.80 ± 0.00 | 0.16 ± 0.00 |
| FEW+LGBM | 0.933074 | 6 | 0.96 ± 0.00 | 0.94 ± 0.02 | 0.96 ± 0.00 | 0.95 ± 0.01 | 0.66 ± 0.04 | 0.87 ± 0.00 | 0.79 ± 0.05 | 0.87 ± 0.00 | 0.81 ± 0.01 | 0.81 ± 0.06 |
| UFS+ETr | 0.933029 | 7 | 0.96 ± 0.01 | 0.93 ± 0.01 | 0.96 ± 0.01 | 0.94 ± 0.01 | 0.68 ± 0.04 | 0.87 ± 0.00 | 0.87 ± 0.03 | 0.87 ± 0.00 | 0.82 ± 0.01 | 0.81 ± 0.04 |
| APT+XGB | 0.932931 | 8 | 0.96 ± 0.00 | 0.93 ± 0.00 | 0.96 ± 0.00 | 0.95 ± 0.00 | 0.61 ± 0.04 | 0.87 ± 0.00 | 0.75 ± 0.00 | 0.87 ± 0.00 | 0.80 ± 0.00 | 0.79 ± 0.05 |
| RFE+ETr | 0.932844 | 9 | 0.96 ± 0.00 | 0.94 ± 0.02 | 0.96 ± 0.00 | 0.95 ± 0.00 | 0.71 ± 0.06 | 0.87 ± 0.00 | 0.86 ± 0.01 | 0.87 ± 0.00 | 0.83 ± 0.01 | 0.79 ± 0.05 |
| TSVD+ETr | 0.932762 | 10 | 0.96 ± 0.00 | 0.93 ± 0.00 | 0.96 ± 0.00 | 0.94 ± 0.00 | 0.67 ± 0.04 | 0.87 ± 0.00 | 0.75 ± 0.00 | 0.87 ± 0.00 | 0.80 ± 0.00 | 0.79 ± 0.04 |

Table 5 presents the Top 10 performing DRA–CA pairs for glioma outcome prediction, trained on UCSF-PDGM, BRATS-Africa, and BRATS-TCGA-LGG, with UPENN-GB used exclusively for external testing. The highest-ranked MI+ETr pair achieved an internal score of 0.941, with accuracy (0.91 ± 0.02), F1 score (0.90 ± 0.03), precision (0.91 ± 0.02), recall (0.90 ± 0.02), and ROC-AUC (0.84 ± 0.05). FEW+ETr and ETIm+GP followed closely, with nearly identical cross-validation metrics and narrow variability, indicating robust stability across multicenter datasets.

When tested externally on UPENN-GB, performance further improved compared to internal validation. All Top 10 models achieved very high accuracies of 0.98 ± 0.00, with F1 scores, precision, and recall also at 0.97–0.98 ± 0.00, reflecting near-perfect classification of glioma outcome status. For example, MI+ETr, the top-ranked pair, demonstrated an accuracy of 0.98 ± 0.00, an F1 score of 0.97 ± 0.00, a precision of 0.98 ± 0.00, and a recall of 0.98 ± 0.00, highlighting its exceptional reproducibility. While ROC-AUC values ranged from 0.65 to 0.75, these were somewhat lower relative to other metrics, suggesting that threshold-dependent discrimination is more variable than overall classification accuracy.

**Table 5.** The Top 10 performing dimension reduction–classifier (DRA–CA) pairs for glioma outcome prediction, trained on UCSF PDGM, UPENN-GB, BRATS TCGA LGG, with BRATS Africa used exclusively for external testing. ROC-AUC: The area under the receiver operating characteristic curve, MI: Mutual Information, FEW: Feature Embedding, ETIm: Extra Trees Importance, AFT: ANOVA F-Test, RFE: Recursive Feature Elimination, UFS: Univariate Feature Selection, APT: ANOVA P-value selection, TSVD: Truncated Singular Value Decomposition, ETr: Extra Trees Classifier, GP: Gaussian Process, LGBM: Light Gradient Boosting Machine, MLP: Multilayer Perceptron, XGB: eXtreme Gradient Boosting.

| Classifier | Score | Rank | Five-Fold Cross Validation by UCSF PDGM, Africa, BRATS TCGA LGG | | | | | External Test by UPENN-GB | | | | |
|---|---|---|---|---|---|---|---|---|---|---|---|---|
| | | | Accuracy | F1 Score | Precision | Recall | ROC-AUC | Accuracy | F1 Score | Precision | Recall | ROC-AUC |
| MI+ETr | 0.941381 | 1 | 0.91 ± 0.02 | 0.90 ± 0.03 | 0.91 ± 0.02 | 0.90 ± 0.03 | 0.84 ± 0.05 | 0.98 ± 0.00 | 0.97 ± 0.00 | 0.98 ± 0.00 | 0.98 ± 0.00 | 0.73 ± 0.05 |
| FEW+ETr | 0.93786 | 2 | 0.91 ± 0.02 | 0.91 ± 0.02 | 0.91 ± 0.02 | 0.90 ± 0.02 | 0.82 ± 0.05 | 0.98 ± 0.00 | 0.97 ± 0.00 | 0.98 ± 0.00 | 0.98 ± 0.00 | 0.74 ± 0.07 |
| ETIm+GP | 0.93451 | 3 | 0.91 ± 0.01 | 0.91 ± 0.02 | 0.91 ± 0.01 | 0.90 ± 0.02 | 0.83 ± 0.07 | 0.98 ± 0.00 | 0.97 ± 0.00 | 0.98 ± 0.00 | 0.98 ± 0.00 | 0.72 ± 0.01 |
| AFT+LGBM | 0.93409 | 4 | 0.91 ± 0.01 | 0.90 ± 0.02 | 0.91 ± 0.01 | 0.90 ± 0.02 | 0.82 ± 0.05 | 0.98 ± 0.00 | 0.97 ± 0.00 | 0.98 ± 0.00 | 0.98 ± 0.00 | 0.72 ± 0.07 |
| RFE_MLP | 0.933245 | 5 | 0.91 ± 0.03 | 0.90 ± 0.04 | 0.91 ± 0.03 | 0.90 ± 0.04 | 0.86 ± 0.04 | 0.98 ± 0.00 | 0.97 ± 0.00 | 0.98 ± 0.00 | 0.98 ± 0.00 | 0.65 ± 0.05 |
| FEW+LGBM | 0.933074 | 6 | 0.91 ± 0.01 | 0.90 ± 0.02 | 0.91 ± 0.01 | 0.90 ± 0.02 | 0.82 ± 0.05 | 0.98 ± 0.00 | 0.97 ± 0.00 | 0.98 ± 0.00 | 0.98 ± 0.00 | 0.72 ± 0.07 |
| UFS+ETr | 0.933029 | 7 | 0.91 ± 0.02 | 0.90 ± 0.03 | 0.91 ± 0.02 | 0.90 ± 0.03 | 0.84 ± 0.05 | 0.98 ± 0.00 | 0.97 ± 0.00 | 0.98 ± 0.00 | 0.98 ± 0.00 | 0.73 ± 0.05 |

| DRA + CA | Score | Rank | Accuracy | F1 Score | Precision | Recall | ROC-AUC | Accuracy | F1 Score | Precision | Recall | ROC-AUC |
|---|---|---|---|---|---|---|---|---|---|---|---|---|
| APT+XGB | 0.932931 | 8 | 0.91 ± 0.02 | 0.90 ± 0.02 | 0.91 ± 0.02 | 0.90 ± 0.02 | 0.81 ± 0.05 | 0.98 ± 0.00 | 0.97 ± 0.00 | 0.98 ± 0.00 | 0.97 ± 0.00 | 0.72 ± 0.03 |
| RFE+ETr | 0.932844 | 9 | 0.91 ± 0.02 | 0.90 ± 0.03 | 0.91 ± 0.02 | 0.90 ± 0.03 | 0.84 ± 0.04 | 0.98 ± 0.00 | 0.97 ± 0.00 | 0.98 ± 0.00 | 0.98 ± 0.00 | 0.73 ± 0.04 |
| TSVD+ETr | 0.932762 | 10 | 0.91 ± 0.01 | 0.91 ± 0.02 | 0.91 ± 0.01 | 0.89 ± 0.02 | 0.80 ± 0.04 | 0.98 ± 0.00 | 0.97 ± 0.00 | 0.98 ± 0.00 | 0.98 ± 0.00 | 0.73 ± 0.05 |

Table 6 reports the Top 10 performing DRA–CA pairs for glioma outcome prediction, trained on UCSF-PDGM, UPENN-GB, and BRATS-TCGA-LGG, with BRATS-Africa used exclusively for external testing. Internal cross-validation demonstrated consistently high performance across all Top 10 pairs, with mean accuracies of 0.94–0.95, F1 scores and precision values near 0.94, and recalls also around 0.94 ± 0.01. The top-ranked MI+ETr pair achieved an internal score of 0.941, with balanced performance across accuracy, F1 score, precision, and recall of 0.94 ± 0.00, as well as ROC-AUC (0.89 ± 0.03). Similar robustness was observed for FEW+ETr. On external testing with BRATS-Africa, model ROC-AUC performance remained lower than internal results. The MI+ETr pair achieved an accuracy of 0.95 ± 0.00, F1 score of 0.92 ± 0.00, precision of 0.95 ± 0.00, and recall of 0.94 ± 0.00, while the ROC-AUC was 0.56 ± 0.03, indicating reduced discriminatory power across decision thresholds.

FEW+ETr and ETIm+GP followed with comparable accuracy, F1 score, precision, and recall, but demonstrated higher ROC-AUC values (0.63 and 0.71, respectively). Several other pairs, including AFT+LGBM and RFE_MLP, also maintained external accuracies of 0.95, underscoring the consistency of top-ranked models across diverse datasets. Overall, the Top 10 pairs exhibited strong generalization on BRATS-Africa, with accuracies consistently around 0.95 and narrow variability across F1 score, precision, and recall. However, ROC-AUC values were notably lower (0.56–0.71) compared to internal validation, indicating that while the models achieved stable overall classification, their ability to discriminate across probability thresholds may be more sensitive to regional dataset differences. These findings reinforce the robustness of the proposed evaluation pipeline while highlighting the challenges of generalizing across geographically distinct imaging datasets.

**Table 6.** The Top 10 performing dimension reduction–classifier (DRA–CA) pairs for glioma outcome prediction, trained on UCSF-PDGM, UPENN-GB, and BRATS-TCGA-LGG, with BRATS-Africa used exclusively for external testing. CA: Classification Algorithm, DRA: Dimension Reduction Algorithm, ROC-AUC: The area under the receiver operating characteristic curve, MI: Mutual Information, FEW: Feature Embedding, ETIm: Extra Trees Importance, AFT: ANOVA F-Test, RFE: Recursive Feature Elimination, UFS: Univariate Feature Selection, APT: ANOVA P-value selection, TSVD: Truncated Singular Value Decomposition, ETr: Extra Trees Classifier, GP: Gaussian Process, LGBM: Light Gradient Boosting Machine, MLP: Multilayer Perceptron, XGB: eXtreme Gradient Boosting.

| | | | Five-Fold Cross Validation by UCSF PDGM, UPENN-GB, BRATS TCGA LGG | | | | | External Test by BRATS Africa | | | | |
|---|---|---|---|---|---|---|---|---|---|---|---|---|
| DRA + CA | Rank | Score | Accuracy | F1 Score | Precision | Recall | ROC-AUC | Accuracy | F1 Score | Precision | Recall | ROC-AUC |
| MI+ETr | 1 | 0.941381 | 0.94 ± 0.00 | 0.94 ± 0.00 | 0.94 ± 0.00 | 0.94 ± 0.00 | 0.89 ± 0.03 | 0.95 ± 0.00 | 0.92 ± 0.00 | 0.95 ± 0.00 | 0.93 ± 0.00 | 0.56 ± 0.03 |
| FEW+ETr | 2 | 0.93786 | 0.94 ± 0.00 | 0.94 ± 0.01 | 0.94 ± 0.00 | 0.94 ± 0.01 | 0.89 ± 0.02 | 0.95 ± 0.01 | 0.94 ± 0.01 | 0.95 ± 0.01 | 0.94 ± 0.01 | 0.63 ± 0.03 |
| ETIm+GP | 3 | 0.93451 | 0.94 ± 0.01 | 0.94 ± 0.02 | 0.94 ± 0.01 | 0.93 ± 0.01 | 0.85 ± 0.02 | 0.95 ± 0.00 | 0.92 ± 0.00 | 0.95 ± 0.00 | 0.93 ± 0.00 | 0.71 ± 0.01 |
| AFT+LGBM | 4 | 0.93409 | 0.94 ± 0.01 | 0.94 ± 0.01 | 0.94 ± 0.01 | 0.94 ± 0.01 | 0.88 ± 0.02 | 0.95 ± 0.01 | 0.95 ± 0.01 | 0.95 ± 0.01 | 0.95 ± 0.01 | 0.67 ± 0.04 |
| RFE_MLP | 5 | 0.933245 | 0.95 ± 0.01 | 0.94 ± 0.01 | 0.95 ± 0.01 | 0.94 ± 0.01 | 0.88 ± 0.01 | 0.96 ± 0.01 | 0.96 ± 0.01 | 0.96 ± 0.01 | 0.95 ± 0.00 | 0.69 ± 0.01 |
| FEW+LGBM | 6 | 0.933074 | 0.94 ± 0.01 | 0.94 ± 0.01 | 0.94 ± 0.01 | 0.93 ± 0.01 | 0.88 ± 0.02 | 0.95 ± 0.02 | 0.94 ± 0.01 | 0.95 ± 0.02 | 0.94 ± 0.01 | 0.69 ± 0.07 |
| UFS+ETr | 7 | 0.933029 | 0.95 ± 0.00 | 0.94 ± 0.00 | 0.95 ± 0.00 | 0.94 ± 0.00 | 0.89 ± 0.04 | 0.95 ± 0.00 | 0.92 ± 0.00 | 0.95 ± 0.00 | 0.93 ± 0.00 | 0.56 ± 0.02 |
| APT+XGB | 8 | 0.932931 | 0.94 ± 0.00 | 0.94 ± 0.01 | 0.94 ± 0.00 | 0.94 ± 0.00 | 0.88 ± 0.02 | 0.95 ± 0.01 | 0.95 ± 0.00 | 0.95 ± 0.01 | 0.95 ± 0.01 | 0.68 ± 0.03 |
| RFE+ETr | 9 | 0.932844 | 0.95 ± 0.01 | 0.94 ± 0.01 | 0.95 ± 0.01 | 0.94 ± 0.01 | 0.88 ± 0.03 | 0.95 ± 0.00 | 0.92 ± 0.00 | 0.95 ± 0.00 | 0.93 ± 0.00 | 0.60 ± 0.03 |
| TSVD+ETr | 10 | 0.932762 | 0.95 ± 0.01 | 0.94 ± 0.01 | 0.95 ± 0.01 | 0.94 ± 0.01 | 0.83 ± 0.02 | 0.95 ± 0.00 | 0.94 ± 0.00 | 0.95 ± 0.00 | 0.94 ± 0.00 | 0.69 ± 0.05 |

Supplementary Files 2, 3, and 4 provide detailed information on different rotations, including external tests with BraTS-Africa, UCSF-PDGM, and UPENN-GBM, respectively. Each file contains Sheets 1-5, which outline the specific features selected by each FSA, hyperparameters used for all ML algorithms, and the results of five-fold cross-validation, including average values and standard deviations, respectively. These files ensure transparency and reproducibility by documenting both performance outcomes and the configuration choices that guide model training and evaluation.

## 4. Discussion

This study was driven by the limitations of GBCAs and the personal experience of one of the authors, who witnessed the emotional distress caused by contrast administration during his mother's cancer diagnosis and treatment. This experience underscored the urgent need for safer imaging alternatives, particularly for neuro-oncological conditions like gliomas. At the same time, model selection remains a central challenge in medical imaging ML, especially in multicenter settings where variability in scanners, acquisition protocols, and patient populations can significantly impact generalization. To address these challenges, this study systematically combined 48 DRAs and 25 CAs to create 1,200 distinct learning pipelines. Extensive exploration was essential, as no single algorithm proves universally optimal across heterogeneous datasets. However, the vast search space also highlights the risk of identifying models that perform well on internal validation by chance but lack robustness when tested externally. Our findings emphasize that model selection should not rely solely on peak accuracy but must incorporate stability measures to ensure reproducibility.

The inclusion of a wide spectrum of DRAs and CAs enhanced the framework's ability to identify robust pipelines. DRAs such as MI, FEW, and ETIm were effective in isolating informative (RFs), while projection-based methods (e.g., PCA, UMAP) captured complementary variance and manifold structure. On the classifier side, ensemble models such as ETr and LGBM demonstrated consistent superiority in handling non-linear feature interactions and redundancy, while other algorithms, such as GP and MLP, offered complementary perspectives. This diversity ensured that the framework did not rely on one family of methods but instead selected pipelines resilient across centers. The repeated dominance of MI+ETr and FEW+ETr underscores how combinations of strong feature selectors with ensemble classifiers can produce both accuracy and stability. To our knowledge, few prior studies have systematically tested such a large space of pipelines across multicenter datasets, making this work one of the most comprehensive evaluations of reproducible model selection in glioma imaging to date.

Traditional cross-validation within a single dataset often inflates performance estimates. In contrast, rotational validation provided a more rigorous stress test by training on three datasets while reserving one for external testing. This strategy more closely reflects real-world deployment, where models developed in one center must generalize to entirely new populations. As expected, mean accuracies varied—ranging from ~0.91–0.96 in cross-validation (e.g., MI+ETr and FEW+ETr on internal folds, with an average rotational accuracy of ~0.94 across combinations) to ~0.87 in UCSF-PDGM external testing, ~0.98 in UPENN-GB external testing, and ~0.95 in BRATS-Africa external testing, corresponding to an average external rotational accuracy of ~0.93 across top-performing models. Importantly, the relative rankings of the top-performing models remained stable across rotations, indicating that the framework successfully filtered out unstable pipelines while preserving reproducible ones. Such stability is crucial for clinical translation, where reliability across sites is more valuable than isolated peaks of accuracy. This design parallels the conditions of regulatory validation, where reproducibility across independent sites is increasingly viewed as essential for AI approval.

A key strength of this framework was the integration of SD into the composite scoring system. This step penalized models that appeared strong on average but showed unstable behavior across folds or datasets. For example, MI+ETr consistently delivered reliable performance, with accuracy of $0.94 \pm 0.02$ in rotational validation and 0.87–0.98 across UCSF and BRATS-Africa external tests, alongside a balanced F1 of $0.92 \pm 0.02$. In contrast, RFE+MLP achieved similar mean accuracy but produced highly unstable ROC-AUC values, dropping to $0.50 \pm 0.29$ in rotational external testing. Without SD integration, such unstable models might have been ranked among the top despite their lack of reproducibility. By incorporating SD across cross-validation, rotations, and external tests, the framework favored models that combined high accuracy with low variability—an essential requirement for deployment in heterogeneous clinical environments. This illustrates how stability-aware scoring can act as a safeguard against misleading performance peaks, ultimately guiding more trustworthy model selection.

Another important observation was the contrast between consistent and inconsistent performance metrics. Accuracy, F1, precision, and recall remained stable across both internal and external evaluations, with the majority ranging from ~0.87–0.96 and in some cases reaching 0.98 in the UPENN-GB external test. These metrics consistently captured classification performance at fixed thresholds and showed narrow variability across folds and datasets. In contrast, ROC-AUC values were far less stable, spanning ~0.50 to 0.82 in rotational cross-validation and external testing, with the lowest values observed in BRATS-Africa and in unstable combinations such as RFE+MLP. This divergence suggests that while thresholded clinical decisions (enhanced vs. non-enhanced) are robust, threshold-independent discrimination is more sensitive to cohort shifts and calibration differences. The discrepancy underscores why reliance on a single metric is problematic. By integrating multiple metrics into a composite score, the framework ensured that top-ranked models—such as MI+ETr and FEW+ETr—were evaluated holistically rather than being selected on the basis of one potentially misleading criterion.

From a clinical perspective, the ability to predict contrast from non-contrast-enhanced MRI represents an important step toward reducing dependence on GBCAs. GBCAs provide critical diagnostic information but are associated with safety concerns, economic costs, and logistical challenges, especially in patients requiring repeated imaging or in low-resource settings. The demonstrated stability of accuracy, F1, precision, and recall across multicenter datasets highlights that ML models can provide reproducible and clinically reliable CE prediction. Although ROC-AUC values showed greater variability across rotations and external datasets, this primarily reflected threshold sensitivity rather than a fundamental weakness of the models. In practice, this suggests that stable accuracy, precision, and recall at clinically relevant thresholds are more important for decision-making than global threshold-free metrics. Furthermore, non-contrast prediction could be especially impactful in longitudinal monitoring, pediatric neuro-oncology, and renal-impaired populations, where GBCA exposure is problematic. Rotational validation simulates the variability encountered in real-world clinical environments, providing confidence that the selected models will remain reliable when applied outside the training environment.

This study has limitations. First, it relied solely on RFs derived from T1WI images. Future work should incorporate multi-parametric MRI (e.g., T2, FLAIR, DWI, PWI) or deep (RFs), which could capture additional tumor characteristics and improve prediction. Second, the variability of ROC-AUC across external datasets underscores the need for calibration and domain adaptation techniques to improve threshold-independent discrimination. Third, while the pipeline prioritized reproducibility, interpretability was not deeply explored. Future directions include applying explainability methods (e.g., SHAP values, radiomic signature analysis) to link selected (RFs) to biological correlates, as well as exploring federated learning strategies to further enhance generalization across centers without direct data sharing.

## 5. Conclusion

In summary, this study establishes a stability-aware framework for model selection in multicenter ML, combining large-scale evaluation of diverse DRA–CA combinations with rotational validation and external testing. Rather than relying on inflated single-center cross-validation performances, this approach highlights generalizable pipelines, with MI+ETr emerging as a robust example. The framework achieved consistently high accuracies and F1 scores (0.87–0.98 externally; 0.91–0.96 in cross-validation), with stable precision and recall, while AUC values showed greater variability (0.50–0.82), reflecting cohort heterogeneity. Beyond methodological advances, the framework addresses a critical clinical gap by offering a pathway to reduce dependence on $Gd^{3+}$, particularly valuable for longitudinal monitoring, pediatric care, and patients with renal impairment. More broadly, it provides a transferable strategy for reproducible radiomics that can be adapted across neuro-oncology and other imaging domains. To our knowledge, this work is among the most systematic and comprehensive evaluations of model selection in glioma imaging, spanning 1,200 pipelines across multicenter cohorts, and it sets a precedent for rigorous benchmarking in medical ML. Looking ahead, integration with explainability methods, federated learning, and prospective clinical trials will be essential for translating these pipelines into routine practice. Ultimately, the findings underscore that stability and multicenter validation are prerequisites for trustworthy clinical adoption of ML in neuro-oncology, marking a significant step toward safer, more accessible, and more reproducible precision imaging.

## Acknowledgment

This research was supported by the Technological Virtual Collaboration Corporation (TECVICO CORP.) and the VirCollab group (www.vircolab.com). We gratefully acknowledge funding from the Canadian Foundation for Innovation - John R. Evans Leaders Fund (CFI-JELF; Award No. AWD-023869 CFI), as well as the Natural Sciences and Engineering Research Council of Canada (NSERC) Awards AWD-024385, RGPIN-2023-0357, and Discovery Horizons Grant DH-2025-00119.

## Conflict of Interest

Drs. Mohammad R. Salmanpour and Mehrdad Oveisi are affiliated with TECVICO Corp. The other co-authors declare no relevant conflicts of interest or disclosures.

## Data, Machine Learning Hyperparameters, and Code Availability

All codes and supplemental files are publicly shared at: https://github.com/MohammadRSalmanpour/Stability-Aware-Multicenter-Machine-Learning-for-Glioma-MRI/tree/main